\documentclass{article}

\usepackage{graphicx}
\usepackage{amsmath}

\begin{document}

\title{Intuitive visualization of the intelligence\\for the run-down of terrorist wire-pullers}

\author{Yoshiharu Maeno\footnote{Corresponding author, Graduate School of Systems Management, Tsukuba University, Otsuka Bunkyo-ku, 112-0012 Tokyo, Japan. email: maeno.yoshiharu@nifty.com.} and Yukio Ohsawa\footnote{School of Engineering, University of Tokyo, Hongo Bunkyo-ku, 113-8656 Tokyo, Japan. email: ohsawa@sys.t.u-tokyo.ac.jp.}}

\maketitle

\begin{abstract}
The investigation of the terrorist attack is a time-critical task. The investigators have a limited time window to diagnose the organizational background of the terrorists, to run down and arrest the wire-pullers, and to take an action to prevent or eradicate the terrorist attack. The intuitive interface to visualize the intelligence data set stimulates the investigators' experience and knowledge, and aids them in decision-making for an immediately effective action. This paper presents a computational method to analyze the intelligence data set on the collective actions of the perpetrators of the attack, and to visualize it into the form of a social network diagram which predicts the positions where the wire-pullers conceals themselves.
\end{abstract}

\section{Introduction}
\label{Introduction}

Terrorist attacks cause great economic, social and environmental impacts. The disaster by the terrorist attacks is different from the emergence arising from earthquakes or hurricanes, in that active non-routine responses are always necessary, and often possible. The investigation of the terrorist attacks and the responses are, however, a time-critical task. The investigators have a limited time window to diagnose the organizational background of the terrorists, to run down and arrest the wire-pullers, and to take an action to prevent or eradicate the attack. The arrest of the wire-pullers is more likely to dismantle the terrorist organization than that of the perpetrators. This is because a limited number of persons can provide the perpetrators with financial supports and elaborate plots, while a perpetrator can be replaced by another person easily. 

Let us show an example in the 9/11 terrorist attack in 2001. Mustafa A. Al-Hisawi, whose alternate name was Mustafa Al-Hawsawi, was alleged to be a wire-puller, who had acted as a financial manager of Al Qaeda. He had attempted to help terrorists enter the United States, and provided the hijackers of the 4 aircrafts with financial support worth more than 300,000 dollars, according to Wikipedia (free encyclopedia on WWW). Furthermore, Usama bin Laden is suspected to be a wire-puller behind Mustafa A. Al-Hisawi and the conspirators behind the hijackers. 

This paper presents a computational method to analyze the intelligence data set on the collective actions of the perpetrators of the terrorist attack, and to visualize it into the form of a social network diagram which predicts the positions where the wire-pullers conceals themselves. The intuitive interface between the intelligence data set and the investigator through the social network diagram is expected to stimulate the investigators' experience and knowledge, and to aid them in decision-making for an immediately effective action. Mathematically, the objective of the analysis is to solve a node discovery problem in a complex network. Two layers are assumed in this problem. The first layer describes the latent pattern of the communication among persons, which is the transmission of the influence on decision-making. It is modeled by a graph structure, where nodes are persons, and links indicate the presence of communication among the persons. The structure is assumed not to be observable directly. The second layer describes the observable pattern of collective actions of the persons. It appears as a result of the communication in the latent layer. Part, or all of this pattern is assumed to be observable. To solve the node discovery problem means to discover some clues on the covert nodes, which participate in the communication in the latent layer, but do not appear in the collective actions in the observable layer.

Related works are reviewed briefly in the section \ref{Related}. A method to solve the node discovery problem is developed in the section \ref{Method}. The section \ref{attack} demonstrates an example where the perpetrators and a wire-puller in the 9/11 terrorist attack are analyzed. The section \ref{Conclusion} concludes this paper.

\section{Related works}
\label{Related}

Social network analysis \cite{Car05} is a study of social structures, which are made of nodes (individuals, organizations etc.), which are linked by one or more specific types of relationship (transmission of influence, presence of trust etc.). Terrorist or criminal organizations have been studied empirically \cite{Sag04}. Factor analysis is applied to study email exchange in Enron, which ended in bankruptcy due to the institutionalized accounting fraud \cite{Kei06}. The criminal organizations tend to be strings of inter-linked small groups that lack a central leader, but to coordinate their activities along logistic trails and through bonds of friends. Hypothesis can be built by paying attention to remarkable white spots and hard-to-fill positions in a network \cite{Kle02}. The conspirators in the 9/11 terrorist organization are relevant in reducing the distance between hijackers, and enhancing communication efficiently \cite{Kre02}. The 9/11 terrorists' social network is investigated from the viewpoint of efficiency and security trade-off \cite{Mor07}. More security-oriented structure arises from longer time-to-task of the terrorists' objectives. The conspirators improve communication efficiency, preserving hijackers' small visibility and exposure.

Research interests have been extending from describing characteristic nature, toward modeling and predicting unknown phenomena. A hidden Markov model and a Bayesian network are combined to predict the behavior of terrorists \cite{Sin04}. A link discovery and node discovery are typical problems in prediction tasks. The link discovery predicts the existence of an unknown link between two nodes from the information on the known attributes of the nodes and the known links \cite{Get05}. The Markov random network is an undirected graphical model similar to a Bayesian network, which is used to learn the dependency between the links which share a node \cite{And99}, \cite{Fra86}. The link discovery techniques are combined with domain-specific heuristics and expertise, and applied to many practical problems. The Markov random network is applied to collaborative classification of web pages \cite{Tas03}. Prediction of the collaboration between scientists from the published co-authorship is studied \cite{Lib07}. Inference of the friendship between people from the information available on their web pages is studied \cite{Ada03}. Missing links in a hierarchical network is predicted by estimating the parameters of a dendrogram, which generates the observed network structure \cite{Cla08}. The dendrogram is a tree diagram used to illustrate the arrangement of the hierarchical clusters.

On the other hand, the node discovery predicts the existence of an unknown node around the known nodes from the information on the collective behavior of the network. Related works in the node discovery is, however, limited. Heuristic method for node discovery is proposed in \cite{Mae07}, \cite{Ohs05}. The method is applied to analyze the covert social network foundation behind the terrorism disasters \cite{Mae08}. Learning techniques of latent variables can be employed, once the presence of a node is known. \cite{Sil06} studied learning of a structure of a linear latent variable graph. \cite{Fri98} studied learning of a structure of a dynamic probabilistic network. But, while the accuracy of the heuristic method is limited, these principled analytic approaches in learning are not practical to handle real human relationship and communication observed in a social network, where much complexity appears. The complexity includes bi-directional and cyclic influence among many observed and latent nodes. We need an efficient and accurate method to solve the node discovery problem. 

\section{Method for inference, discovery, and visualization}
\label{Method}

\subsection{Intelligence data set}
\label{Intelligence}

\begin{figure}
\begin{center}
\includegraphics[scale=0.45,angle=-90]{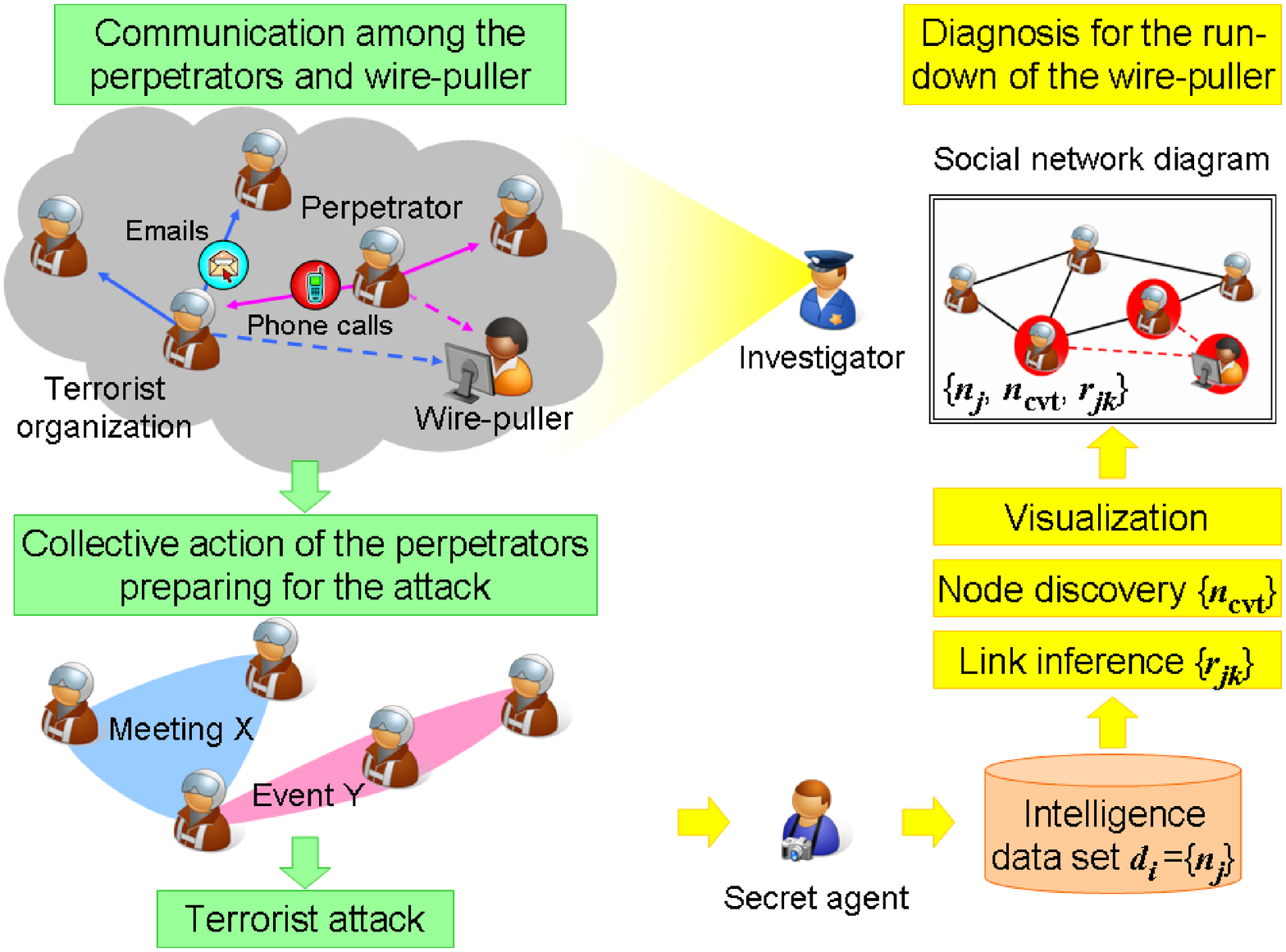}
\end{center}
\caption{Investigator diagnosing the intelligence data set for the run-down of the wire-puller behind the terrorist attack. The communication and influence on decision-making among the perpetrators and a wire-puller in the terrorist organization governs the pattern of the collective actions (such as attending an meeting or event) of the perpetrators who have prepared for the attack. The communication and influence are assumed not to be observable directly. Part or all of the collective actions are assumed to be observable. The intelligence data set consists of the secret agents' records on the collective actions of the perpetrators. The wire-puller does not appear in the intelligence. After link inference and node discovery, the intelligence data set is visualized into an intuitively understandable social network diagram.}
\label{ivirtw1}
\end{figure}

Imagine a situation where an investigator diagnoses the intelligence data set for the run-down of the wire-puller behind the terrorist attack. Figure \ref{ivirtw1} illustrates the situation. The pattern of the communication among perpetrators and a wire-puller in the terrorist organization lies in the latent layer. It is the transmission of the influence on decision-making. The pattern governs that of the collective actions of the perpetrators who have prepared for the terrorist attack in the observable layer. The collective actions are attending an meeting or event, for example. 

The intelligence data set consists of the secret agents' records on the collective actions of the perpetrators. The wire-puller, however, does not appear in the intelligence. The wire-puller is referred to by a covert node $n_{{\rm cvt}}$. The investigator applies three tools: (1) link inference to infer the communication between the perpetrators, (2) node discovery to discover the position of the wire-puller, and (3) visualization of the result of the inference and discovery into the form of a social network diagram. The social network diagram serves as an intuitive interface to understand the intelligence data set. 

Mathematical definition of the intelligence data set is given here. In eq.(\ref{delta}), $\delta_{i}$ represents the participants to single communication in the latent layer. It is a set of nodes $n_{j}$.

\begin{equation}
\delta_{i} = \{ n_{j} \}.
\label{delta}
\end{equation}

It is assumed that the individual intelligence data obtained from the observable layer is the pattern in eq.(\ref{delta}) minus the covert node $n_{{\rm cvt}}$. It is denoted by eq.(\ref{Dataset2}).

\begin{equation}
d_{i} = \delta_{i} \backslash \{ n_{{\rm cvt}} \}.
\label{Dataset2}
\end{equation}

The whole intelligence data set is denoted by $D$ in eq.(\ref{Dataset1}). The number of intelligence data is $|d|$.

\begin{equation}
D=\{d_{i}\} \ (0 \leq i \leq |d|-1).
\label{Dataset1}
\end{equation}

The intelligence data set $D$ can be expressed by a two dimensional matrix of binary variables. The presence, or absence of the node $n_{j}$ in the intelligence data $d_{i}$ is represented by eq.(\ref{BasketVector}). The number of node species is given by $|n|$.

\begin{equation}
d_{ij} = \left \{ \begin{array}{ll}
                    1 & \mbox{\ \ if $n_{j} \in d_{i}$} \\
                    0 & \mbox{\ \ otherwise}
                \end{array}
         \right . (0 \leq j \leq |n|-1) .
\label{BasketVector}
\end{equation}

The intelligence data set is the input to the method for link inference, node discovery, and visualization. The nodes in an intelligence data do not necessarily form a clique structure, where there are links between every pair of nodes. Assuming that they formed a clique would result in a very densely connected structure in the latent layer. Such a superficial interpretation of the intelligence data set leads to a wrong understanding of the terrorist organization. This is why we need a new computational method. The method is described below.

\subsection{Link inference}
\label{Link}

The communication in the latent layer is modeled by a graph. Strictly speaking, a graph, a diagram, and a map are defined differently. A graph is an abstract mathematical entity, which consists of a set whose elements are nodes, and a set of pairs of nodes. A diagram is a 2 dimensional expression of a graph, which is the drawing of the web of the links interconnecting at the nodes. In a map, faces are relevant, which are determined by the links and their intersections. A network refers to either a graph, or a diagram. The map is not used in this paper. The nodes in the graph represent persons, and the links represent the transmission of influence. The links between the nodes which appeared in the intelligence data set is inferred with the maximum likelihood estimation (MLE). A parametric form is defined to describe the link topology. The probability where the influence transmits from an initiating node $n_{j}$ to a responder node $n_{k}$ is $r_{jk}$. The influence transmits to multiple responders independently in parallel. It is similar to the degree of collaboration probability in trust modeling \cite{Lav07}. Eq.(\ref{CondR1}) gives constraints on $r_{jk}$.

\begin{equation}
0 \leq r_{jk} \leq 1, \ r_{jj} = 1 \ (0 \leq j, k \leq |n|-1).
\label{CondR1}
\end{equation}

The quantity $f_{j}$ is the probability where the node $n_{j}$ becomes an initiator. The parameters $r_{jk}$ and $f_{j}$ are denoted by \mbox{\boldmath{$r$}} collectively. It is the target to estimate from the observation data set.

\begin{equation}
\mbox{\boldmath{$r$}} = \{ r_{jk} \}  \cup \{ f_{j} \} \ (0 \leq i, j \leq |n|-1).
\label{ParameterSet}
\end{equation}

The logarithmic likelihood function \cite{Dud00} is defined by eq.(\ref{Likelihood1}). In statistics, a likelihood function is a conditional probability function of the observation given the parameters of a statistical model. It plays a key role in statistical inference such as Bayes' Law. The probability where $D$ occurs for given \mbox{\boldmath{$r$}} is denoted by $p(D|\mbox{\boldmath{$r$}})$.

\begin{equation}
L(\mbox{\boldmath{$r$}}) = \log( p(D|\mbox{\boldmath{$r$}})).
\label{Likelihood1}
\end{equation}

The individual observations are assumed to be independent. Eq.(\ref{Likelihood1}) becomes eq.(\ref{Likelihood2}).

\begin{equation}
L(\mbox{\boldmath{$r$}}) = \log( \prod_{i=0}^{|d|-1} p(d_{i}|\mbox{\boldmath{$r$}})) = \sum_{i=0}^{|d|-1} \log( p(d_{i}|\mbox{\boldmath{$r$}})).
\label{Likelihood2}
\end{equation}

The quantity $f_{k|ij}$ in eq.(\ref{fijk1}) is the probability where the presence or absence of the node $n_{k}$ as a responder to the stimulating node $n_{j}$ coincides with the observation $d_{i}$. 

\begin{equation}
f_{k|ij} = \left \{ \begin{array}{ll}
                    r_{jk} & \mbox{\ \ if $d_{ik} = 1$ for given $i$ and $j$} \\
                    1-r_{jk} & \mbox{\ \ otherwise}
                \end{array}
         \right ..
\label{fijk1}
\end{equation}

Eq.(\ref{fijk1}) is equivalent to eq.(\ref{fijk2}) since the value of $d_{ik}$ is either $0$ or $1$.

\begin{equation}
f_{k|ij} = d_{ik} r_{jk} + (1-d_{ik})(1-r_{jk}).
\label{fijk2}
\end{equation}

The probability $p(D|\mbox{\boldmath{$r$}})$ in eq.(\ref{Likelihood2}) is expressed by eq.(\ref{Prob1}).

\begin{equation}
p(d_{i}|\mbox{\boldmath{$r$}}) = \sum_{j=0}^{|n|-1} d_{ij} f_{j} \prod_{{\scriptsize \begin{array}{c} 0 \leq k \leq |n|-1 \\ k \neq j \end{array} }} f_{k|ij}.
\label{Prob1}
\end{equation}

The logarithmic likelihood function is expressed by eq.(\ref{Likelihood3}).

\begin{equation}
L(\mbox{\boldmath{$r$}}) = \sum_{i=0}^{|d|-1} \log ( \sum_{j=0}^{|n|-1} d_{ij} f_{j} \prod_{k=0}^{|n|-1} \{ 1-d_{ik}+(2d_{ik}-1)r_{jk} \} ).
\label{Likelihood3}
\end{equation}

The maximal likelihood estimator $\hat{\mbox{\boldmath{$r$}}}$ is obtained by solving eq.(\ref{Estimator}). It represents the inferred topology of the network.

\begin{equation}
\hat{\mbox{\boldmath{$r$}}} = \arg \max_{\mbox{\boldmath{$r$}}} L(\mbox{\boldmath{$r$}}).
\label{Estimator}
\end{equation}

Lagrange multipliers can be used to solve eq.(\ref{Estimator}) analytically. But, at present, computational optimization is suitable to solve a large-scale problem. The hill climbing method is a simple incremental optimization technique. Unsuitable selection of the initial condition may lead to the sub-optimal solutions. Advanced meta-heuristic algorithms such as simulated annealing, or genetic algorithm \cite{Has01} may be employed to avoid sub-optimal solutions. It is not within the scope of this paper to explore the computational technique to solve eq.(\ref{Estimator}). The details of the algorithm implementation are not described here.

\subsection{Node Discovery}
\label{Node}

The clues on the covert node in the latent layer are discovered after the topology of the links between the nodes, which appeared in the intelligence data set, is inferred with the maximum likelihood estimation (MLE) in \ref{Link}. The degree of suspiciousness ($s(d_{i})$) is assigned to the individual intelligence data $d_{i}$. It is defined as the likeliness where the covert node would appear in the intelligence data, if it became overt, or if the wire-puller were observable. The degree of suspiciousness is evaluated by eq.(\ref{Ranking1}), where $g(x)$ is a monotonically decreasing function of the variable $x$. Larger value in eq.(\ref{Ranking1}) means more suspicious intelligence data. 

\begin{equation}
s(d_{i}) \propto g(p(d_{i}|\hat{\mbox{\boldmath{$r$}}})).
\label{Ranking1}
\end{equation}

Ranking of the intelligence data can be calculated from the value of eq.(\ref{Ranking1}). The $k$-th most suspicious intelligence data is given by $d_{\sigma(k)}$ in eq.(\ref{Ranking2}). The degree of suspiciousness of $d_{\sigma(j)}$ ($s(d_{\sigma(j)})$) is larger than that of $d_{\sigma(k)}$ ($s(d_{\sigma(k)})$) for any $j<k$. 

\begin{equation}
\sigma(k) = \arg \max_{{\scriptsize \begin{array}{c} i \neq \sigma(j) \\ j<k \end{array} }} s(d_{i}) \ (0 \leq k \leq |d|-1).
\label{Ranking2}
\end{equation}

The degree of suspiciousness ($s(n_{j})$) can also be assigned to the individual nodes $n_{j}$. More suspicious node is more likely to be the neighbor node of the covert node. Or, it is more likely to be the perpetrator who is associated with the wire-puller closely. The degree of suspiciousness $s(n_{j})$ can be evaluated by accumulating the degree of suspiciousness of the intelligence data ($s(d_{i})$), where the node appears, as in eq.(\ref{Ranking3}). The function $w(k)$ is an appropriate weight function. 

\begin{equation}
s(n_{j}) \propto \frac{ \sum_{k=0}^{|d|-1} w(k) B(n_{j} \in d_{\sigma(k)}) }{ \sum_{k=0}^{|d|-1} B(n_{j} \in d_{k}) }.
\label{Ranking3}
\end{equation}

The Boolean function in eq.(\ref{Ranking3}) is defined by eq.(\ref{boolean}).

\begin{equation}
B(x) = \left \{ \begin{array}{ll}
          1 & \mbox{if proposition $x$ is True} \\
          0 & \mbox{otherwise}
        \end{array}
    \right ..
\label{boolean}
\end{equation}

Alternatively, much simpler means to discover the suspicious nodes may be taken. The suspicious nodes are the neighbor nodes $n_{{\rm nbr}}$, which appear in the most suspicious intelligence data $d_{\sigma(0)}$. It is denoted by eq.(\ref{Ranking4}).

\begin{equation}
\{ n_{{\rm nbr}} \} \in d_{\sigma(0)}.
\label{Ranking4}
\end{equation}

\subsection{Social network visualization}
\label{Visualization}

The output of the link inference and node discovery is drawn as a social network diagram, which predicts the position of the covert node $n_{{\rm cvt}}$ with the suspicious neighbor nodes $n_{{\rm nbr}}$. The link topology is relevant in the social network diagram. But, the absolute position of the nodes, the distance between the nodes, the direction along the vertical and horizontal axes are not relevant. The position of the nodes is determined by the employed graph drawing algorithm. For example, the spring model \cite{Fru91} is popular. The model converts the strength of the relationship across the link between two nodes into Hooke's constant of the spring, which is placed between the nodes imaginarily, and calculates the equilibrium position of the nodes.

At first, the social network of the nodes which appeared in the intelligence data set is drawn. The communication in the latent layer is present between the nodes $n_{j}$ and $n_{k}$, if the maximal likelihood estimator $\hat{\mbox{$r$}}_{jk}$ is larger than $0$. The estimator $\hat{\mbox{$r$}}_{jk}$ approaches to either 1 or 0, in most cases. The link is drawn if $\hat{\mbox{$r$}}_{jk} > r_{{\rm thr}}$, because strong relationship is of interest to the investigator. The threshold $r_{{\rm thr}}$ may be 0.9 or larger. Next, the suspicious nodes $n_{{\rm nbr}}$ in eq.(\ref{Ranking4}) are added into the social network diagram with the discovered covert node $n_{{\rm cvt}}$. The identity of the covert node $n_{{\rm CVT}}$ is not known at this stage of investigation. The covert node is, therefore, represented by an unlabeled (unnamed) node. The links between the covert node and the suspicious neighbor nodes are drawn. 

The parameter $\hat{\mbox{$f$}}_{j}$, which is the probability where the node $n_{j}$ becomes an initiator of communication, is not visualized into the topology of the social network diagram. If it is necessary to indicate the value graphically, the largeness of the value can be encoded into the color, or largeness of the symbol of the node.

\section{Visualization of the 9/11 terrorists}
\label{attack}

\begin{table}

\caption{The 19 perpetrators who are responsible for hijacking the 4 commercial flights (number: AA11, AA77, UA93, and UA175) in the 9/11 terrorist attack, and appear in a sample intelligence data set.}
\label{table1}
\begin{center}
\begin{tabular}{|l|l|}
\hline
Flight & Hijacker \\
\hline
\hline
AA11 & Abdul A Al-Omari, Mohamed Atta, Satam Suqami, Wail Alshehri,\\
 & Waleed Alshehri\\
\hline
AA77 & Hani Hanjour, Khalid Al-Mihdhar, Majed Moqed, Nawaf Alhazmi,\\
 & Salern Alhazmi\\
\hline
UA93 & Ahmed Al-Haznawi, Ahmed Alnami, Saeed Alghamdi, Ziad Jarrah\\
\hline
UA175 & Ahmed Alghamdi, Fayez Ahmed, Hamza Alghamdi, Marwan Al-Shehhi,\\
 & Mohand Alshehri\\
\hline
\end{tabular}
\end{center}

\end{table}

\begin{table}

\caption{Sample intelligence data set on the perpetrators in Table \ref{table1}, which is the input to the method (eq.(\ref{Dataset1}) in \ref{Intelligence}) to generate a social network diagram.}
\label{table2}
\begin{center}
\begin{tabular}{|l|l|}
\hline
Data & Participant to a collective action \\
\hline
\hline
$d_{0}$ & \{Abdul A Al-Omari, Fayez Ahmed, Hani Hanjour, Marwan Al-Shehhi,\\
 & Mohamed Atta, Salern Alhazmi, Ziad Jarrah\} \\
\hline
$d_{1}$ & \{Abdul A Al-Omari, Hani Hanjour, Marwan Al-Shehhi, Mohamed Atta,\\
 & Nawaf Alhazmi, Ziad Jarrah\} \\
\hline
$d_{2}$ & \{Abdul A Al-Omari, Marwan Al-Shehhi, Mohamed Atta, Waleed Alshehri\} \\
\hline
$d_{3}$ & \{Abdul A Al-Omari, Satam Suqami, Wail Alshehri, Waleed Alshehri\} \\
\hline
$d_{4}$ & \{Ahmed Alghamdi, Ahmed Al-Haznawi, Ahmed Alnami, Hamza Alghamdi,\\
 & Mohand Alshehri, Nawaf Alhazmi, Saeed Alghamdi\} \\
\hline
$d_{5}$ & \{Ahmed Alghamdi, Hamza Alghamdi\} \\
\hline
$d_{6}$ & \{Ahmed Al-Haznawi, Ahmed Alnami, Hamza Alghamdi, Nawaf Alhazmi,\\
 & Saeed Alghamdi\} \\
\hline
$d_{7}$ & \{Ahmed Al-Haznawi, Hamza Alghamdi, Saeed Alghamdi, Ziad Jarrah\} \\
\hline
$d_{8}$ & \{Ahmed Al-Haznawi, Hani Hanjour, Marwan Al-Shehhi, Mohamed Atta,\\
 & Salern Alhazmi, Ziad Jarrah\} \\
\hline
$d_{9}$ & \{Ahmed Alnami, Hamza Alghamdi, Hani Hanjour, Khalid Al-Mihdhar,\\
 & Mohamed Atta, Nawaf Alhazmi, Saeed Alghamdi, Salern Alhazmi\} \\
\hline
$d_{10}$ & \{Ahmed Alnami, Hamza Alghamdi, Nawaf Alhazmi, Saeed Alghamdi\} \\
\hline
$d_{11}$ & \{Fayez Ahmed, Hamza Alghamdi, Mohand Alshehri\} \\
\hline
$d_{12}$ & \{Fayez Ahmed, Marwan Al-Shehhi, Mohamed Atta, Waleed Alshehri\} \\
\hline
$d_{13}$ & \{Fayez Ahmed, Marwan Al-Shehhi, Mohand Alshehri\} \\
\hline
$d_{14}$ & \{Hani Hanjour, Khalid Al-Mihdhar, Majed Moqed, Marwan Al-Shehhi,\\
 & Mohamed Atta, Nawaf Alhazmi, Ziad Jarrah\} \\
\hline
$d_{15}$ & \{Hani Hanjour, Khalid Al-Mihdhar, Nawaf Alhazmi\} \\
\hline
$d_{16}$ & \{Hani Hanjour, Majed Moqed\} \\
\hline
$d_{17}$ & \{Marwan Al-Shehhi, Nawaf Alhazmi, Salern Alhazmi, Ziad Jarrah\} \\
\hline
$d_{18}$ & \{Satam Suqami, Wail Alshehri, Waleed Alshehri\} \\
\hline
$d_{19}$ & \{Satam Suqami, Wail Alshehri, Waleed Alshehri\} \\
\hline
\end{tabular}
\end{center}

\end{table}

\begin{figure}
\begin{center}
\includegraphics[scale=0.45,angle=-90]{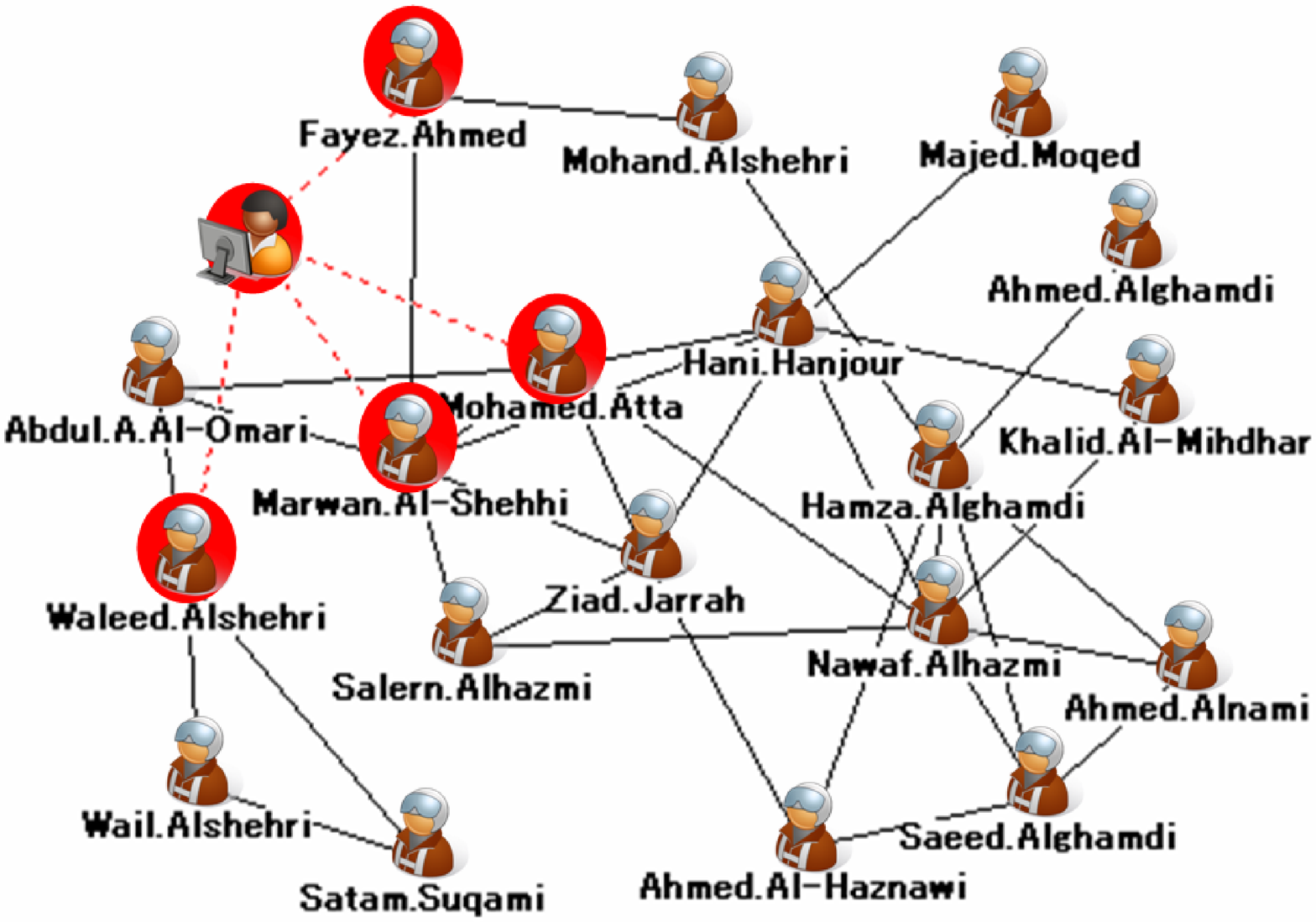}
\end{center}
\caption{Social network diagram which visualizes the links between the nodes, and the node discovered from the sample intelligence data set in Table \ref{table2}. The nodes represent the 19 perpetrators. The links represent the transmission of influence between the perpetrators, which governs the pattern of the collective actions which were recorded in the intelligence data set. The discovered node, which is not labeled in the figure, indicates the position where the wire-puller is most likely to conceal himself. The node has 4 links (red broken lines) to Fayez Ahmed, Mohamed Atta, Marwan Al-Shehhi, and Waleed Alshehri. They are the keystone neighbors for the run-down and arrest of the wire-puller. The wire-puller in the figure turns out to be Mustafa A. Al-Hisawi in \cite{Kre02}.}
\label{ivirtw4}
\end{figure}

An example of visualizing the intelligence data set is demonstrated. The intelligence data set is the records on the collective actions of the perpetrators of the 9/11 terrorist attack in 2001. The 19 perpetrators are listed in Table \ref{table1}, who are responsible for hijacking the 4 commercial flights in the 9/11 terrorist attack (number: American Airlines AA11 (Boeing 767 from Boston to Los Angeles), AA77 (Boeing 757 from Washington to Los Angeles), United Airlines AA175 (Boeing 767 from Boston to Los Angeles), and UA93 (Boeing 757 from Newark to San Francisco)), and appear in a sample intelligence data set.

The intelligence data set is listed in Table \ref{table2}. The data set is not real, but is generated for the purpose of computer simulation. The intelligence data set is the input to the method to generate a social network diagram, which predicts the positions where the wire-pullers conceals themselves. The resulting social network diagram is shown in Figure \ref{ivirtw4}. It visualizes the inferred links between the nodes, and the node discovered from the sample intelligence data set in Table \ref{table2}. The nodes represent the 19 perpetrators. The links represent the transmission of influence between the perpetrators in the latent layer. They govern the pattern of the collective actions in the observable layer, which were recorded in the intelligence data set. The perpetrators in an intelligence data do not form a simple clique structure, actually. For example, if the 8 nodes in $d_{9}$ formed a clique structure, the nodal degree of the nodes would be 7. The actual nodal degree is much smaller than 7. The network is much less densely connected than it would be if the perpetrators in a single intelligence data formed a clique structure. The method reveals the less densely connected topology of the terrorist organization.

The discovered node $n_{{\rm cvt}}$, which is not labeled (named) in the figure, indicates the position where the wire-puller is most likely to conceal himself. The node has links (red broken lines) to the 4 neighbor nodes: Fayez Ahmed, Mohamed Atta, Marwan Al-Shehhi, and Waleed Alshehri. The neighbor nodes are $n_{{\rm nbr}} \in d_{12}$, because $d_{\sigma(0)} = d_{12}$ is the most suspicious intelligence data. Mohamed Atta and Waleed Alshehri hijacked the flight AA11, which flew into the North Tower of the World Trade Center. Mohamed Atta was the leader of the hijackers. Waleed Alshehri assisted Mohammed Atta. Fayez Ahmed and Marwan Al-Shehhi hijacked the flight UA175, which flew into the South Tower. Marwan Al-Shehhi, who trained at a pilot school with Mohammed Atta, flew the aircraft. Fayez Ahmed was also a pilot. The wire-puller was a linkage between the 2 flights which were used as weapon to the financial landmark of the United States. They are the keystone neighbors to investigate to unravel who is the wire-puller, or what is behind the terrorist attack.

The ties between the persons concerned are revealed by \cite{Kre02}. They include the 19 perpetrators, conspirators, and suspects. The tie strength is estimated by the amount of time together by a pair of terrorists. Those living together, attending the same school, or the same training would have strong ties. Those traveling together, and participating in meetings together would have ties of moderate strength. Finally, those who were recorded as having a financial transaction together, or participating in an occasional meeting would have weak ties. Publicly available data is used as such intelligence data set. By comparing with the ties in \cite{Kre02}, the wire-puller in Figure \ref{ivirtw4} turns out to be Mustafa A. Al-Hisawi. He is believed to be a financial manager of Al Qaeda, and to give Fayez Ahmed a credit card as a financial support for the attack.

Remind that this result is for demonstration purpose. The objective is to demonstrate how the intuitive visualization can help the investigator diagnose the intelligence data set, rather than to uncover the real role of Mustafa A. Al-Hisawi, or Usama bin Laden in the 9/11 terrorist attack.

\section{Conclusion}
\label{Conclusion}

This paper presented a computational method to analyze the intelligence data set on the collective actions of the perpetrators of the terrorist attack, and to visualize it into the form of a social network diagram, which predicts the positions where the wire-pullers conceals themselves. This will help the investigators, who have a limited time window to diagnose the organizational background of the terrorists, to run down and arrest the wire-pullers, and to take an action to prevent or eradicate the terrorist attack. The intuitive interface to visualize the intelligence data set will stimulate the investigators' experience and knowledge, and aids them in decision-making for an immediately effective action.

The important aspect of the presented method is that the diagnosis of the investigator illustrated in Figure \ref{ivirtw1} can be repeated many times. The investigator can predict the position of Mustafa A. Al-Hisawi from the intelligence data set on the 19 perpetrators. Then, once the wire-puller is identified as Mustafa A. Al-Hisawi, the intelligence data set on Mustafa A. Al-Hisawi should be collected and added to the diagnosis. Similarly, the investigator can predict the position of the 21st person again from the intelligence data set on the 19 perpetrators and Mustafa A. Al-Hisawi.

The method provides the investigator with the intuitively comprehensible direction of potentially fruitful investigation from what is already known toward what is not, but can be known.

\end{document}